\documentclass[twocolumn]{article}

\usepackage{authblk}

\usepackage{algorithm}
\usepackage{algorithmic}

\usepackage{times}
\usepackage{helvet}
\usepackage{courier}
\usepackage[hyphens]{url} 
\usepackage{graphicx}

\usepackage{cite}
\usepackage{amsmath,amssymb,amsfonts}
\usepackage{algorithmic}
\usepackage{graphicx}
\usepackage{textcomp}

\usepackage{url}
\usepackage{amsmath}
\usepackage{ascmac}
\usepackage{amsfonts}
\usepackage{latexsym}

\newcommand{\ve}[1]{\mathbf{#1}}

\title{Variational Bayes Decomposition for Inverse Estimation \\with Superimposed Multispectral Intensity}
\author[1]{Akinori Asahara}
\author[1]{Yoshihiro Osakabe}
\author[1]{Mitsuya Yamamoto}
\author[1]{Hidekazu Morita}

\affil[1]{Hitachi Ltd.}

\begin{document}

\maketitle

\begin{abstract}
A variational Bayesian inference for measured wave intensity, such as X-ray intensity, is proposed in this paper.
The data is popular to obtain information about unobservable features of an object, such as a material sample and the components of it.
The proposed method assumes particles represent the wave, and their behaviors are stochastically modeled.
The inference is accurate even if the data is noisy because of a smooth prior setting.
Moreover, in this paper, two experimental results show feasibility of the proposed method.
\end{abstract}

\section{Introduction}

Machine learning has been applied to scientific activities, such as extracting information from observational data\cite{5754705}. 
One example of those is experimental data analysis of microscope images of new materials.
Conventionally, material researchers carefully inspect experimental data to find features in them.
The researchers, however, might miss these features as doing otherwise takes a lot of effort.
Therefore, machine learning to automatically extract features or hints for discovery from experimental data is highly needed.

This paper focuses on unsupervised machine learning applied to inverse estimation with an observed a SMI (Superimposed Multispectral Intensity) of wave.
The observation is often performed to measure directly unobservable features. 
For instance, SAS (Small Angle Scattering) is one of the popular observation methods.
In SAS, a wave ray, such as X-rays, is irradiated into a material sample to obtain the micro grain size of it.
The ray incident upon the sample interacts with the micro grain therein, and the beam is scattered as multispectral rays due to the interactions with the grains.  
Therefore, the scattered-ray intensity reflects the grain size.
However, since the material sample contains grains of multiple sizes, the rays scattered by the grains are superimposed into an intensity observation.
The observed SMI must be decomposed by grain size to explain the sample composition.
Because a component corresponding to a grain size is known in the theory about the scattering phenomena, the decomposition makes the microstructure clear.
Therefore machine learning is applied to the decomposition\cite{9638297}\cite{do2020small}.
Not only limited to SAS, SMI is often obtained for physics observatory such as microscopes, and also there are various multispectral particles, such as from X-ray scattering, ion-beam scattering, etc.
Therefore, a general method of SMI decomposition is expected to be applied to versatile measurements. 

There are existing methods for decomposition, but a method that focuses on robustness is notably lacking.
Reasonable methods are parameter fitting and stochastic inference, such as the ML (Maximum Likelihood) and MAP (Maximum A Posteriori) inference, are known. 
However, the model for these methods can be fit with noise, resulting in quite noisy results: this is called overfitting.
To address this, regularization parameters are often induced into the model, but the parameters are manually adjusted.
There is a risk that the adjustment may introduce arbitrariness in the experimental results.

Therefore, a stochastic method resistant to overfitting is proposed in this work.
The proposed method is based-on VB (Variational Bayes) inference,
where not only observations are stochastic but also parameters.
A prior distribution of the parameters is first assumed in the proposed method.
The prior distribution is revised with observations and the expectation values of the revised distribution lead to the decomposed factors. 
Because the prior distribution restricts the revised distribution, overfitting can be reduced.

\section{Related Work}

There are known multivariable decomposition methods.
PCA (Principal Component Analysis\cite{jolliffe2002principal}) is one of the most popular methods.
PCA extracts the dominant components of given multivariable data and the represents the data as a linear combination of the components.
ICA (Independent Component decomposition Analysis)\cite{fastICA_paper} is another known algorithm to extract components independent from each other.
Components extracted with PCA and ICA can be negative, but in SMI, the coefficient of superposition should not be negative. 
A method called NMF (Non-negative Matrix Factorization)\cite{4408452} decomposes into non-negative factors.
When the basis is not known in component decomposition like PCA, ICA, and NMF, estimation of the basis requires a large number of samples, such as time series data and image data.
They are often called empirical mode decomposition (EMD) \cite{2010rehmanmultivariate}, used for the purpose of denoising and inference of various measurements,
such as tomography\cite{CONG201559}\cite{10136197}\cite{Panontin2021}\cite{9099880}.
These methods estimate both the components and the coefficients. In the SMI inverse estimation problem,
however, the components are known, and the coefficients are to be estimated.

The coefficients optimization is generally based on the least errors.
The decomposition is generally regarded as a linear transformation of observation. Therefore, a matrix representing the transformation can be assumed.
SVD (Singular Value Decomposition\cite{SVD}) gives the pseudoinverse matrix of the transformation matrix.
By multiplying it to the observation, the optimal coefficients are derived.
However, the results of the SVD may include negative values, and also, the calculation is noise sensitive.
Therefore, the parameters should be optimized with constraint. 
Actually, for SAS, IFT (Indirect Fourier Transformation \cite{IFT}), which is a method based on coefficient optimization for an automatic grain-size estimation, is known.
As mentioned above, regularization is added to the optimization to suppress overfitting. The regularization should be adjusted objectively to reduce arbitrariness\cite{Svergun}. However, it is determined retrospectively for lack of reasons. This can be a loss of objectivity.

Stochastic inference is also applicable to the decomposition.
One of the popular stochastic inferences is known as ML inference with EM (Expectation-Maximization) algorithm\cite{asahara2020em}.
The process of observation is modeled by parametric PDF (Probability Distribution Function) and the likelihood is maximized with changing the PDF.
ML inference tends to fit to the noise caused by observation, i.e., overfitting is not yet solved.

MAP inference is another stochastic method that is based on Bayesian statistics, applied to inverse estimation\cite{Lucy}, image reconstruction\cite{895918}, and so on. 
In Bayesian statistics, the parameters of the PDF also obey their PDF (called priors), and they are revised with observations (called posteriors).
The parameters that give the maximum posterior are adopted as the inference result.
The prior, which provides suppression of overfitting, should be prospectively determinable to avoid loss of objectivity.

The method proposed in this paper is Bayesian inference focused on the issue.
In the proposed method, the prior can be set uniform for overfitting reduction without arbitrariness.
Generally, parametric PDFs such as Gaussian are applied to unsupervised learning on Bayesian inference.
However, for the SMI, because the PDFs are complicated in many cases, the prior is difficult to set. 
Addressing the problem, VB with non-parametric PDFs is proposed.
 
\section{Variational Bayes Inference for SMI}

\subsection{Problem Settings}

\begin{figure}[tb]
\centering
\includegraphics[width=0.35\textwidth]{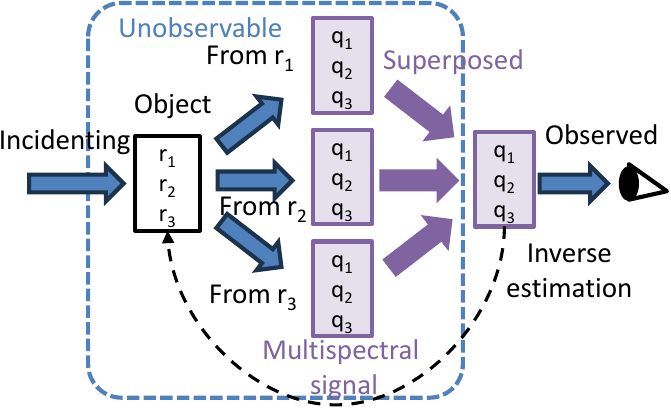}
\caption{Inverse Estimation of SMI }\label{inv_smi}
\end{figure}

In observation process, an incident ray interacts with the object as shown in Fig. \ref{inv_smi}.
The object has multiple components (e.g., mixture of materials), labeled $r=r_i$. Hence, the interaction with each component makes a responding ray. 
The responding rays are superimposed before observation, and observed at each frequency, which is proportional to the wave energy.
Instead of the frequency, a wave number $q$ is used, which is $ 2 \pi$ multiplied by the frequency. 
Accordingly, the responding ray intensity is denoted by a function $I(r, q)$.

The responding rays are superimposed as the weighted sum of $I(r,q)$ over $r$.
Hence, the SMI $S(q)$ is written as 
\begin{align}
S(q) \propto \sum_i f(r_i) I(q,r_i), \label{S_cal}
\end{align}
where the superposition weight distribution is denoted as $f(r)$.
To estimate $f(r)$, $S(q)$ should be decomposed to the summation of $I(q,r)$.
What should be estimated is $f(r)$, which indicates the ratio of the components of the sample.
Note that the observation can include the intensity fluctuation.
For example, when the light ray is incident, the intensity fluctuates due to the quantum mechanical behavior of photons.
If the average over a long period of time is taken, the fluctuation is reduced but it takes a very long time.
Accordingly, noise-resistant analysis is expected to implement efficient measurement.

The VB inference procedures adapted to SMI are proposed in this paper.
As the basic concept of the proposed method, the SMI is modeled as the particle detection counts.
For instance, X-Ray is modeled as a group of photons and the intensity is handled as the number of the photons.
The particle behavior is stochastically modeled to infer the interacted components.

\subsection{Formulation}

VB inference\cite{PatternRecog} is well known as a method of unsupervised machine learnings.
In the VB inference, an approximated formula for PDFs is assumed, and they are revised to converge to the true PDFs.
For the convergence, Kullback-Leibler divergence from the true PDF to the approximated one is minimized.
For example, the following equations are calculated in the process of VB inference to infer PDFs with a parameter $\theta$ and a latent variable $Z$.  
\begin{align}
\ln \tilde P(\theta) = \mathbb{E}_{Z} (\ln P(\theta, Z)) + const,\\
\ln \tilde P(Z) = \mathbb{E}_{\theta} (\ln P(\theta, Z)) + const, \label{VB}
\end{align}
where the total PDF is denoted as $P(\theta, Z)$, and $\tilde P(\cdot)$ represents an approximated PDF.
Note that $\mathbb{E}_{\mathrm{something}} (\cdot)$ indicates an expectation value, which involves integral of a function by $\mathrm{something}$,
and execution of the integration generally requires many computational resources. 
$\tilde P(\theta)$ is derived from the first equation, and $\tilde P(Z)$ is with the second one.

\begin{figure}
\centering
\includegraphics[width=0.3\textwidth]{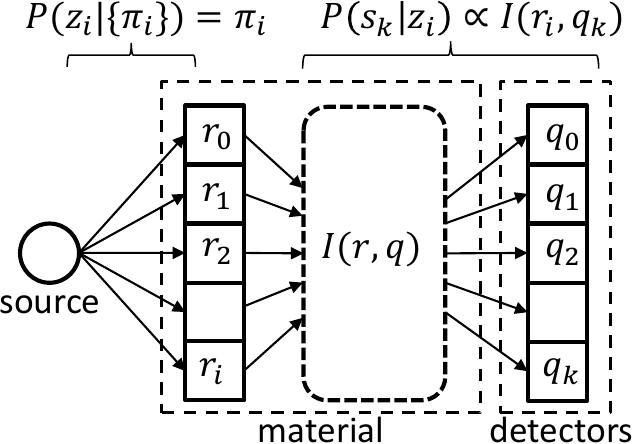}
\caption{Stochastic model illustration}\label{stmodel}
\end{figure}

In the proposed VB inference, the SMI is modeled as the relative frequency of the particle detection, and the stochastic model of the particle behavior is inferred.
Figure \ref{stmodel} illustrates the formulation.
The detected particle is labeled with $n$, and $s_n$ is the wavenumber of the particle labeled $n$, as shown in the figure.
Posterior $P(\{\pi_i\}|\{s_n\})$ is written with the Bayes theorem as follows.
\begin{align}
P(\{\pi_i\}|\{s_n\})  = P(\{s_n\}|\{\pi_i\})P(\{\pi_i\})/P(\{s_n\}),
\end{align}
where $\pi_i$ indicates relative frequency based on $f(r_i)$.
Note that $P(\{s_n\})$ and $P(\{\pi_i\})$ shown in the formula are priors for $\{s_n\}$ and $\{\pi_i\}$.
Therefore, they can be tuned to control the smoothness of the inference result.
What is to be revised with the observation $\{s_n\}$ is $P(\{\pi_i\})$ in VB inference.

As a latent variable during the observation process, the interacting component of the particle is defined as $\{\ve{z}_n\}$.
$\ve{z}_n$ is a vector of which components correspond to $\{r_i\}$.
$z_{ni}=1$ if the $i$th particle is interacting component $r_i$, and $z_{ni}=0$ if not.
Accordingly, the VB inference equations are as follows.  
\begin{eqnarray}
\ln \tilde P(\{\pi_i\}) = \mathbb{E}_{\{\ve{z}_n\}} (\ln P(\{\pi_i\}, \{\ve{z}_n\}, \{s_n\})) +\mathrm{const}, \label{vbsas_1}\\
\ln \tilde P(\ve{z}_n) = \mathbb{E}_{\{\pi_i\}} (\ln P(\{\pi_i\}, \{\ve{z}_n\}, \{s_n\})) +\mathrm{const}. \label{vbsas}
\end{eqnarray}
As a result, $P(\{\pi_i\}|\{s_n\}) = \tilde P(\pi_i)$ is obtained.

\subsection{Algorithm}

\subsubsection{Parameter optimization}
$\ln P(\{\pi_i\}, \{\ve{z}_n\}, \{s_n\})$ in (\ref{vbsas_1}) is separated into three terms in the following formula.
\begin{align}
&\ln P(\{\pi_i\}, \{q_n\}, \{\ve{z}_n\}) \nonumber\\
&= \ln P(\{\pi_i\})+\ln P(\{\ve{z}_n\}|\{\pi_i\}) \nonumber +\ln P(\{q_n\}|\{\ve{z}_n\}).
\end{align}

The first term is the prior of the sample parameters.
Dirichlet distribution is adequate for it:
\begin{eqnarray}
\ln P(\{\pi_i\})=\ln \mathrm{Dir}(\{\pi_i\}|\ve{\alpha}_0)= \sum_i \alpha_{0i} \ln \pi_i,
\end{eqnarray}
where $\ve{\alpha}_0$ is the hyperparameter of the Dirichlet distribution.

The second term models particle interaction choices of $r_i$, where the rate is in proportion to $\pi_i$:
\begin{align}
\ln P(\{\ve{z}_n\}|\{\pi_i\})=\sum_{i,n} z_{ni} \ln \pi_i+\mathrm{const}.
\end{align}
Note that $z_{ni}$ is 1 when $r_i$ is chosen, otherwise zero. 

The third term represents the stochastic process $I(q, r_i)$ to represent that a particle interacted at $r_i$ is detected at wavenumber $q$.
Therefore,  
\begin{align}
\ln P(\{q_n\}|\{\ve{z}_n\})=\sum_{n,i} z_{n,i} \ln I(q_n, r_i)+\mathrm{const}.
\end{align}

By combining these formulas, (\ref{vbsas_1}) is transformed to the following equation.
\begin{eqnarray*}
\ln \tilde P(\pi_i) = \ln P(\{\pi_i\})+\mathbb{E}_{\{\ve{z}_n\}}(\ln P(\{\ve{z}_n\}|\{\pi_i\}))+\mathrm{const}.
\end{eqnarray*}
The second term in this formula can be calculated.
\begin{align}
\mathbb{E}_{\{\ve{z}_n\}} (\ln P(\{\ve{z}_n\}|\{\pi_i\}))=\sum_{n,i} \mathbb{E}(z_{ni})\ln \pi_i.
\end{align}
For simplification, $\rho_{n,i}$ defined as $\rho_{n,i} \equiv \mathbb{E}(z_{ni})$ leads to
\begin{eqnarray}
\ln \tilde P(\pi_i)= \ln \mathrm{Dir} (\{\pi_i\}; \{\sum_n\rho_{ni}+\alpha_{0i}\}) +\mathrm{const}.\label{mstep}
\end{eqnarray}
This formula shows the revision of the hyperparameter $\alpha_0$ with $\rho_{ni}$s.

\subsubsection{Latent variables inference}
Similarly, (\ref{vbsas}) is transformed.
\begin{eqnarray}
\ln \tilde P(\ve{z}_n)=\mathbb{E}_{\{\pi_i\}} (\ln P(\{\ve{z}_n\}|\{\pi_i\}))\nonumber \\
+\ln P(\{q_n\}|\{\ve{z}_n\})+\mathrm{const}
\end{eqnarray}
Each component of the equation is calculated as follows.
\begin{eqnarray}
\mathbb{E}_{\{\pi_i\}} (\ln P(\{\ve{z}_n\}|\{\pi_i\}))=\sum_{n,i} z_{ni} \mathbb{E}_{\pi_i} (\ln \pi_i). \\
\ln P(\{q_n\}|\{\ve{z}_n\}) = \sum_{n,i} z_{ni} I(q_i, r_i).
\end{eqnarray}
$\mathbb{E}_{\pi_i} (\ln \pi_i)$ appears in the formula. This is the expectation value of $\ln \pi_i$ for the Dirichlet distribution with hyperparameter $\alpha_{n}$, well known as
\begin{align}
\mathbb{E}_{\pi_i} (\ln \pi_i)= \Psi(\alpha_i)-\Psi(\sum_i \alpha_i),
\end{align}
where $\Psi(\cdot)$ is digamma function, which is one of the computable special functions.
Therefore, (\ref{vbsas}) is successfully transformed into the following computable form.
\begin{eqnarray}
\ln \tilde P(\ve{z}_n)=\sum_{in} z_{ni} (\mathbb{E}_{\pi_i} (\ln \pi_i)+\ln I(q_i, r_i))+\mathrm{const}. \label{estep}
\end{eqnarray}

\subsubsection{Overall algorithm}
By consolidating (\ref{mstep}) and (\ref{estep}), steps for inference are derived.
$\rho_{ni}$ in (\ref{mstep}) is simply derived from $\tilde P(z_{n,i})$ because $0\times P(z_{ni}=0)$ is zero.
\begin{align}
\rho_{ni} = \tilde P(z_{ni}=1). \label{rho_calc}
\end{align}
Hence, (\ref{estep}) and (\ref{rho_calc}) lead to
\begin{eqnarray}
\ln \rho_{ni} 
=\Psi(\alpha_i)-\Psi(\sum_j \alpha_j)+\ln I(q_n, r_i)+\mathrm{const}.\label{rho}
\end{eqnarray}
The hyperparameter $\alpha_i$ is required for the calculation.
Because (\ref{mstep}) is revision of hyperparameters $\alpha$ in the Dirichlet distribution as discussed above,
\begin{align}
\alpha_i = \sum_n\rho_{ni}+\alpha_{0i}. \label{alpha}
\end{align}
Consequently, the iteration of (\ref{rho}) and (\ref{alpha}) leads to the hyperparameters of the inferred posteriors.

Furthermore, the calculation can be simplified for the implementation of VB inference.
In SMI, which is the number of detection events at each wavenumber,
a group of the detected particles with common wavenumber can be relabeled.
If the representative wavenumbers are set as $q_k$, the labels of $\rho_{ni}$ are also changed.
\begin{align}
\ln \rho_{ki}=  \Psi(\alpha_k)-\Psi(\sum_k \alpha_k)+\ln I(q_k, r_i).
\end{align}
With these formulas, the same components in (\ref{alpha}) are aggregated as
\begin{align}
\alpha_i = \sum_k n_k \frac{\rho_{ki}}{\sum_k \rho_{ki}} +\alpha_{0i}.
\end{align}
$n_k$ is the number of particles with $q_k$, that is, the component of SMI.
Consequently, the algorithm is shown in Algorithm\ref{alg2}.
Note that the hyperparameter $\alpha_0$ appears in $P(\{\pi_i\})$, added to the intensity $n_k$.
$\alpha_0$ equalizes the SMI, controlling noise sensitivity.

\begin{algorithm}[t]
\caption{VB inference for SMI} \label{alg2} 
\centering \footnotesize
\begin{algorithmic} 
\REQUIRE SMI $n_k\geq0$, wavenumber $q_k \geq 0$ $(k=0, 1, \cdots , K)$, 
the components $r_l \geq 0$ where $(l=0,1,\cdots , L)$
\ENSURE $\{ \alpha_i \}$
\STATE $\{ \eta_{l,k} \} \Leftarrow \{ \frac{I(r_l, q_k)}{\sum_m I(r_l, q_k)} \}$, $\alpha_{k} \Leftarrow \sum_l n_k \eta_{l,k}$
\REPEAT
\STATE $\rho_{ki} \Leftarrow \exp \left(\Psi(\alpha_k)-\Psi(\sum_k \alpha_k)+\ln I(q_k, r_i)\right)$
\STATE $\alpha_i \Leftarrow \sum_k n_k \frac{\rho_{ki}}{\sum_k \rho_{ki}}$
\UNTIL{convergence}
\end{algorithmic}
\end{algorithm}

\section{Experiments}

\subsection{Experiment 1: SAS (Small Angle Scattering) }

\subsubsection{Overview}

\begin{figure}
\centering \footnotesize
\includegraphics[width=0.3\textwidth]{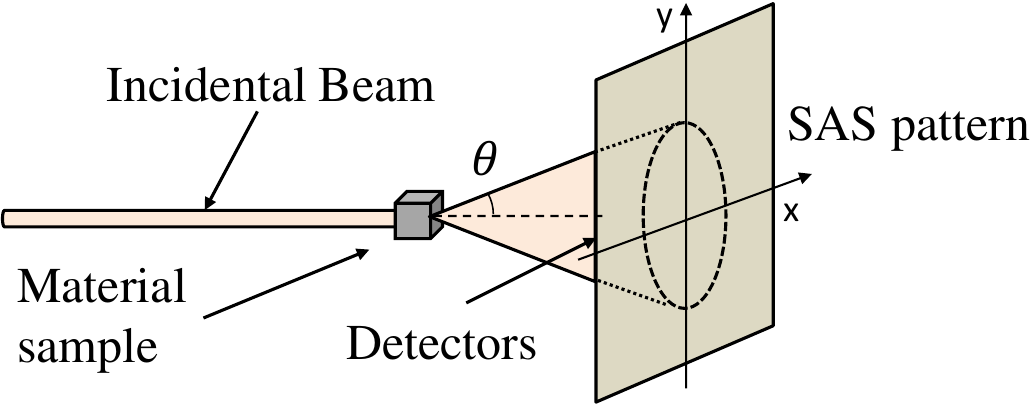}
\caption{SAS experiment set-up }\label{sans_exp}
\end{figure}

Experiment 1 focuses on SAS experiments \cite{higgins1994polymers}.
In the SAS experiment, a particle beam incident upon the sample interacts with the microstructures therein, and the scattered ray is observed as shown in Fig. \ref{sans_exp}.
The simplest grains are assumed to be spheres.
Under this assumption, only the angle $\theta$ between the straight beam and the changed direction of the scattered ray characterizes the interaction.
Detectors arranged in a plane to detect the scattered ray. The counts of the detection events are obtained as the intensity, forming the SMI (called a SAS pattern).

The intensity $I(r, q)$ of a SAS pattern scattered by balls of radius $r$ is in proportion to the following formula:
\begin{align}
I(q,r) \propto \frac{1}{r^3}\left( \frac{\sin qr}{q^3} - \frac{r\cos qr}{q^2}\right)^2. \label{I_form}
\end{align}
The $q$ in the formula indicates the wave number of the scattered ray.
Because the scattering angle $\theta$ is approximately in proportion to $q$, we can obtain the SAS pattern $I(r,q)$ as a pattern on the detector plane.

Several methods for automatic estimation of grain-size distributions using SAS patterns have been proposed\cite{clementi2011bayesian}\cite{hansen1991comparison}\cite{do2020small}\cite{Svergun}, and well-known software\cite{SASView}\cite{SASFit}\cite{ATSAS}\cite{9638297} implements them.
IFT is one of such methods.
For IFT, the formula of the SAS pattern is assumed to be a summation of several stepwise functions, which are reformed as a linear combination of $a_n$.
By minimizing the difference between the linear combination of $a_n$ and the actual SAS pattern, the best $a_n$s are obtained.
In addition, as a stochastic method, ML and MAP inference applied to SAS \cite{asahara2021bayesian} is shown.

If the resolution of the grain size distribution is set high, the setting causes overfitting in these methods.
One technique to avoid this problem is to add regularization terms to suppress the overfitting.
However, it is not easy to adjust the regularization coefficients in advance because it depends on the SAS pattern.
To automate the regularization, methods for determining regularization terms \cite{vestergaard2006application} have been proposed.
However, these methods are based on the analysis of the estimated results. That is, they are consequentialist.

\subsubsection{Experimental settings}
In Experiment 1, simulation-generated SAS pattern datasets were processed so that we could compare the results with the ground truth.
The data were processed with VB inference, the IFT, and ML-based inference\cite{asahara2020em} for comparison.
10,000 iterations were run instead of checking for convergence to simulate a situation where the processing time is limited.
The hyperparameters $\alpha_{0i}$ in VB were set as $\alpha_{0i}=1$, and the regularization parameters of IFT were manually tuned with noiseless data in advance.

Three types of different grain size distributions were defined for the experiments.
Each pattern was the sum of two gamma distributions with the most frequent point around 20nm.
The grain size distribution was discretized by 0.2 nm, and its domain is set from 0 to 60 nm (i.e., 300 values) for $f(r)$ in (\ref{S_cal}).

To obtain the SAS patterns, random samplings were carried out. The detection event number was set as 50,000, and SAS patterns of the grain size distributions were generated.
First, $q$'s domain, which was from 0.1 $\mathrm{nm}^{-1}$ to 5$\mathrm{nm}^{-1}$, was discretized into 200 lots denoted by $q_k$, and $S(q_k)$ was calculated by evaluating the integration of (\ref{S_cal}).
Random sampling along $S(q_k)$, that is, the probability of the detection, was performed to simulate particle detection events and the number of events was counted to generate each SAS pattern.

\subsubsection{Results}

\begin{figure}[tb]
\centering \footnotesize
\begin{tabular}{cc}
\includegraphics[width=0.18\textwidth]{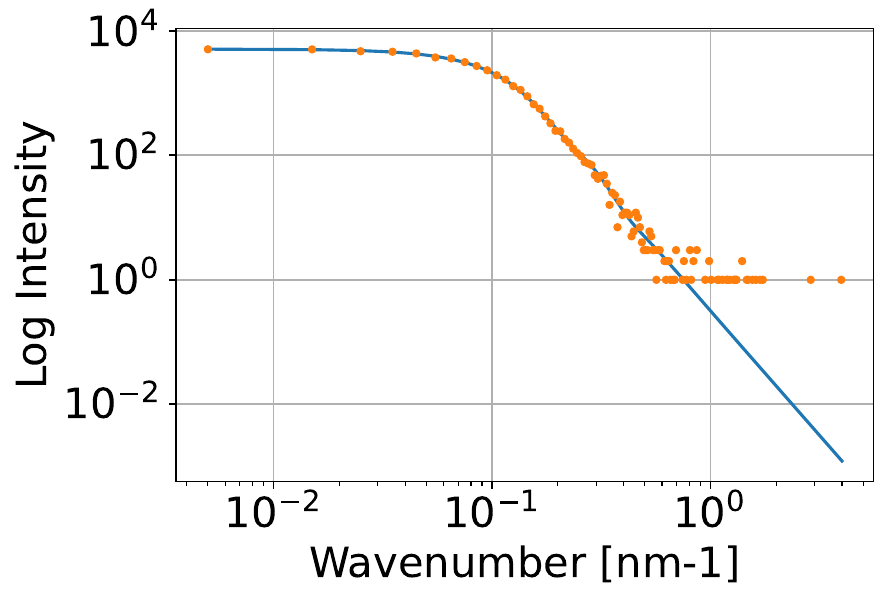} & \includegraphics[width=0.18\textwidth]{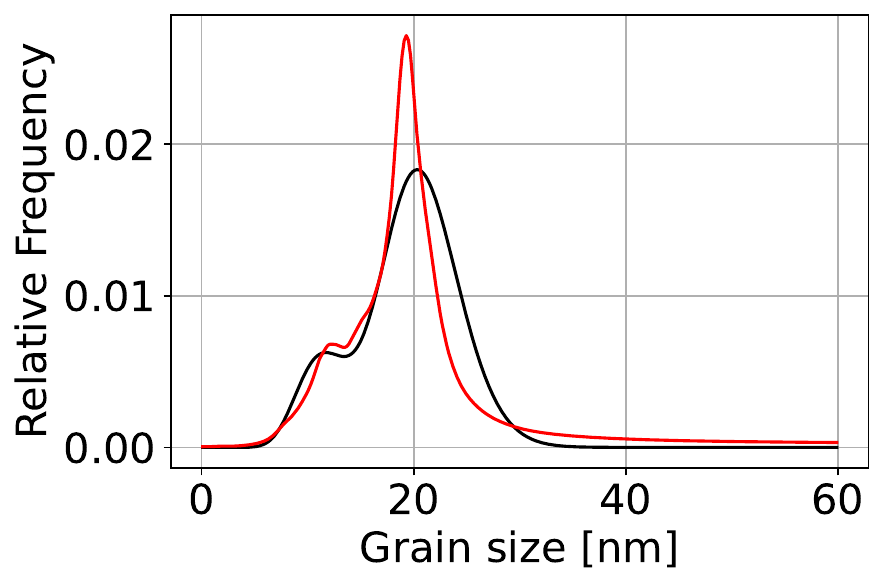} \\
SAS pattern & VB (Proposed) \\
&\\
\includegraphics[width=0.2\textwidth]{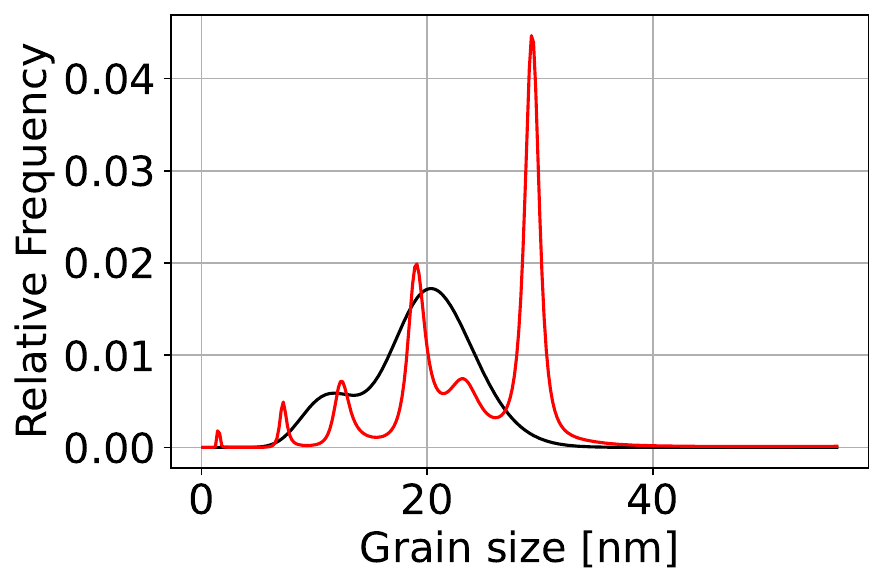} & \includegraphics[width=0.2\textwidth]{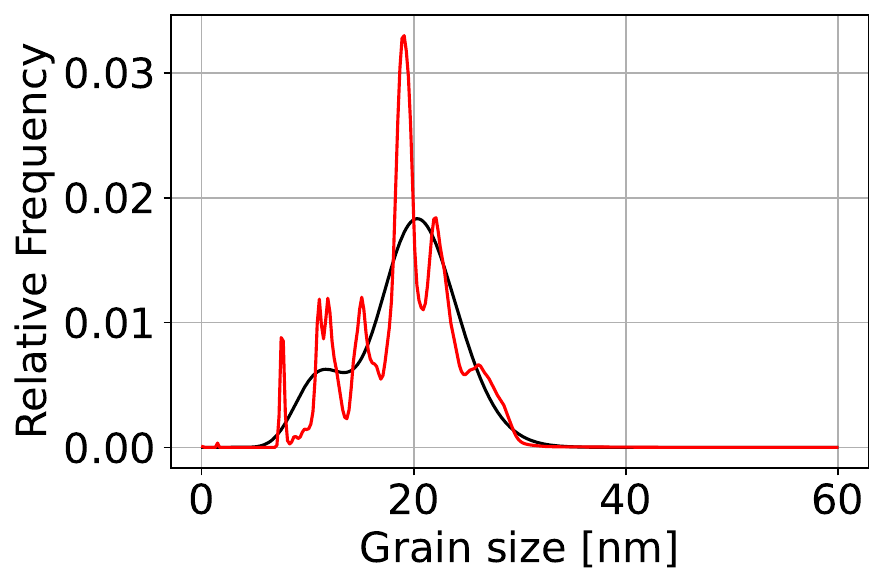}\\
IFT&ML\\
\end{tabular}
\caption{Plateau Pattern inference} \label{exp1_results1}
\end{figure}

\begin{figure}[tb]
\centering \footnotesize
\begin{tabular}{cc}
\includegraphics[width=0.2\textwidth]{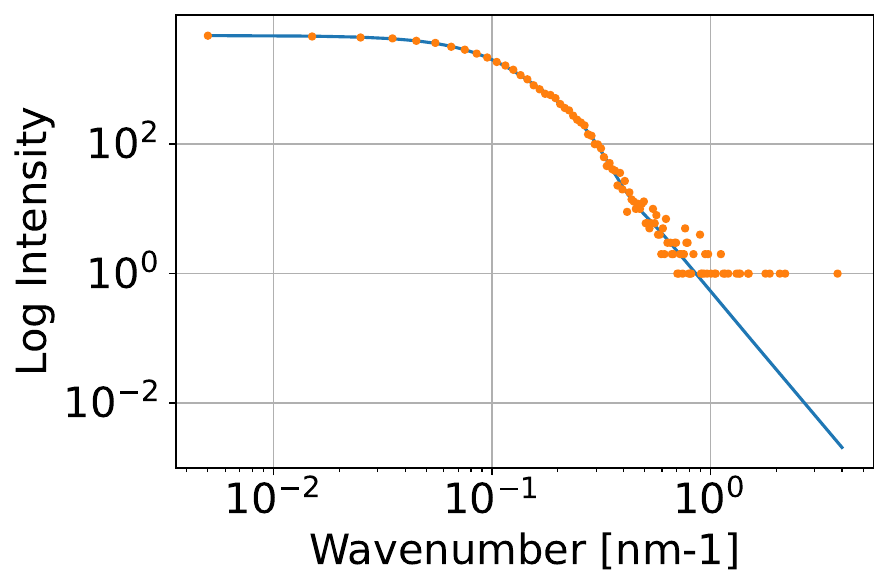} & \includegraphics[width=0.2\textwidth]{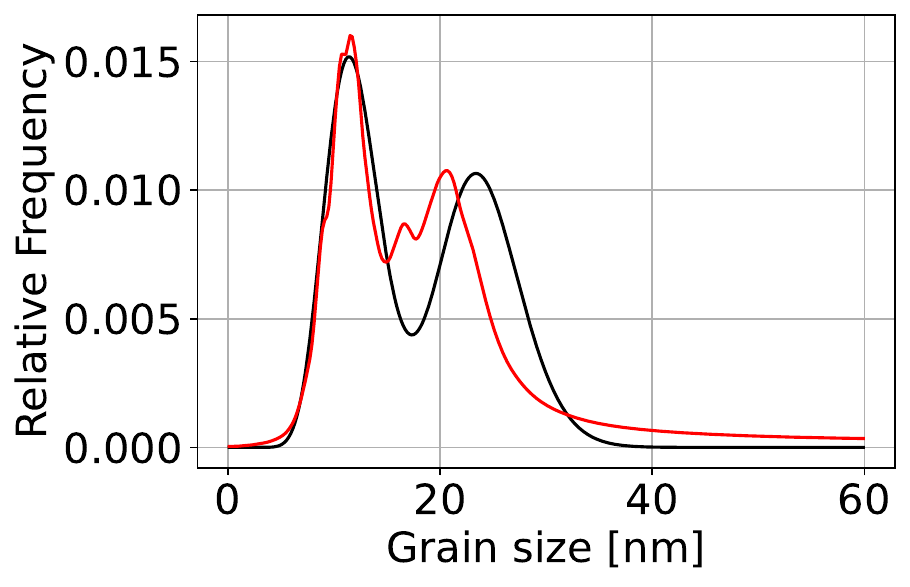} \\
SAS pattern & VB (Proposed) \\
&\\
\includegraphics[width=0.2\textwidth]{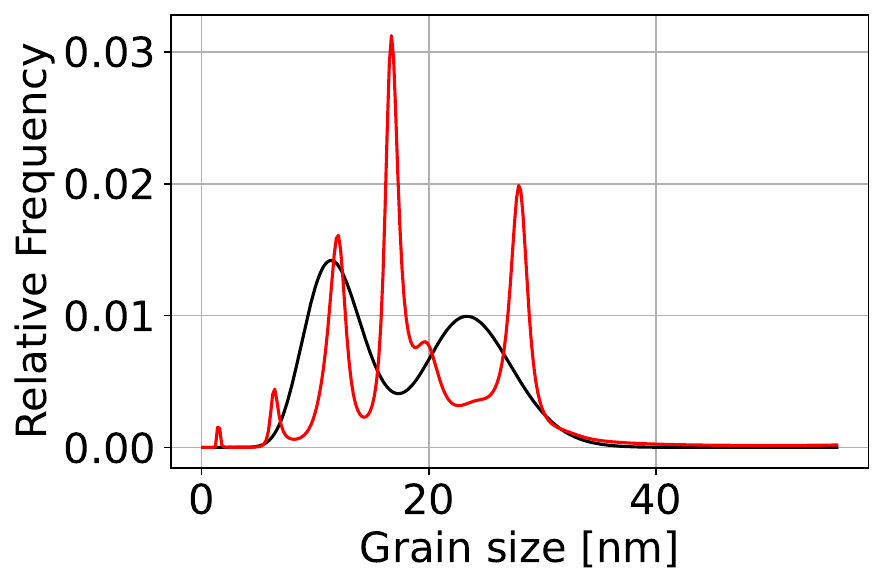} & \includegraphics[width=0.2\textwidth]{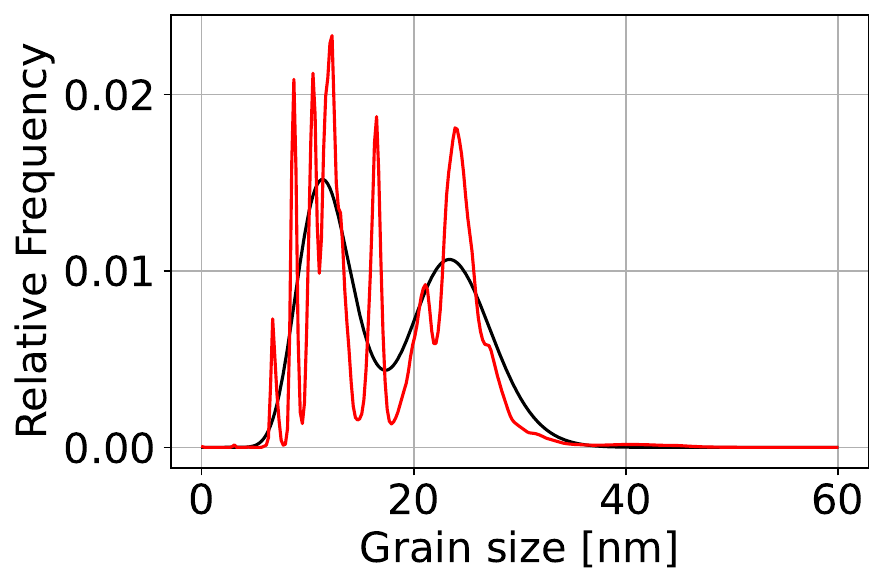}\\
IFT&ML\\
\end{tabular}
\caption{Two-peak pettern inference} \label{exp1_results2}
\end{figure}

\begin{figure}[tb]
\centering \footnotesize
\begin{tabular}{cc}
\includegraphics[width=0.2\textwidth]{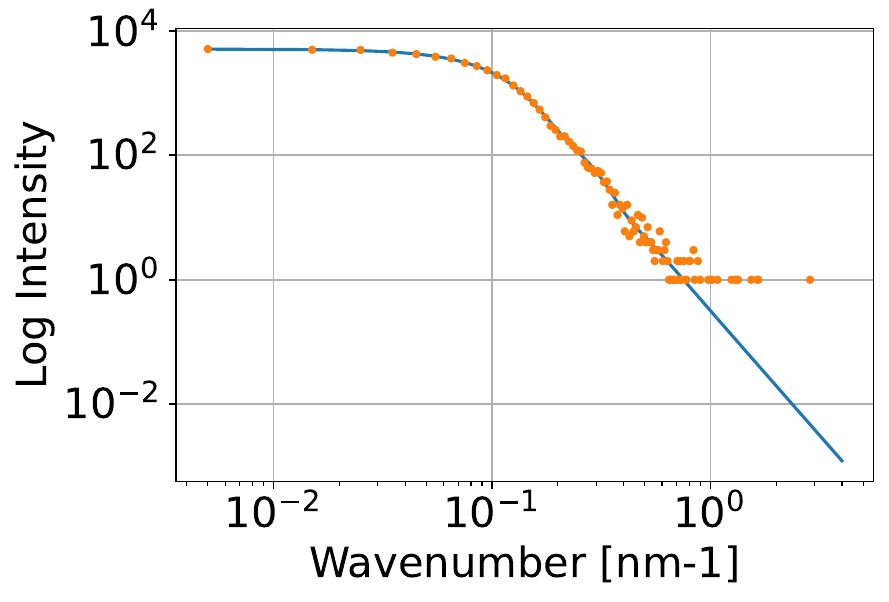} & \includegraphics[width=0.2\textwidth]{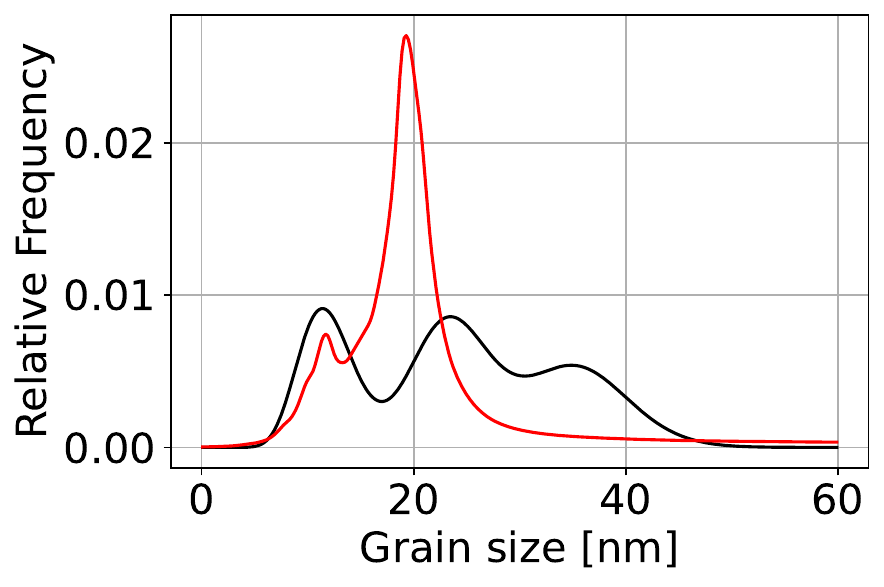} \\
SAS pattern & VB (Proposed) \\
&\\
\includegraphics[width=0.2\textwidth]{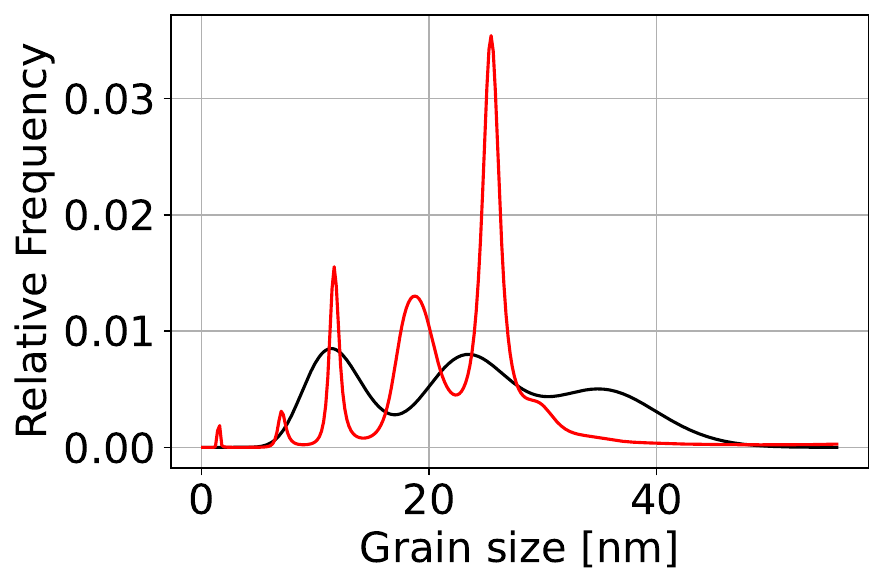} & \includegraphics[width=0.2\textwidth]{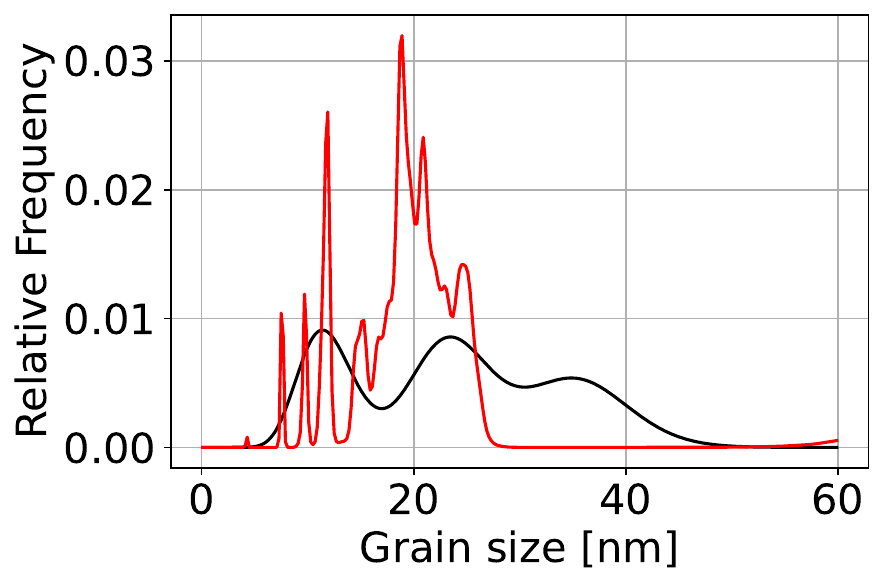}\\
IFT&ML\\
\end{tabular}
\caption{Three-peak pattern inference} \label{exp1_results3}
\end{figure}

Figures \ref{exp1_results1}, \ref{exp1_results2}, and \ref{exp1_results3} show the results. 
The SAS pattern is represented by the log-log plot in the figures.
The blue curves for the SAS pattern are plots for $50,000 \times S(q_k)$, and the orange dots show the SAS patterns generated with $S(q_k)$.
The other plots show the inferred grain size distributions. 
The red lines show the inference results. The result of the VB inference is shown in the plot labeled by ``VB.'' The results of the IFT inference and that of the ML inference are also plotted for comparison.
The black curves in the plots represent the truth, i.e., the original grain size distribution used to generate the SAS pattern.

The first grain size distribution in Fig. \ref{exp1_results1} had only one peak at the center of the $q$ range, and a plateau existed at bottom of the peak.
IFT and ML results were very noisy, but the peak was readable in these results; however, the plateau was difficult to find due to the noise.
In contrast, the VB result was smooth and similar to the truth. The accuracy was sufficient enough to see both the peak and the plateau. 

The second pattern had two peaks, shown in Fig. \ref{exp1_results2}. 
Similarly to the previous pattern, the IFT and ML results were noisy, and the VB result was smooth.
The two peaks were not seen in the IFT and ML results. However, the VB result was outstandingly clear to read the two peaks.

The third grain size distribution, presented in Fig. \ref{exp1_results3}, had three peaks.
The basic trend of the graphs was similar to the others, that is, the IFT and the ML result were too noisy to see the three peaks.
This indicates that a more complicated microstructure is more indistinct in noise.
The VB inference also did not find the three peaks but two peaks. It is the better result even though the microstructures to be inferred were complicated.

\subsubsection{Discussion}

All VB results were similar to the truth, and the other results had noise.
In the ML results, the noise in the small $r$ region was larger than in the other regions.
This indicates that the noise came from the smallness of the sample size.
Remember that the $I(q,r)$ formula indicates that a high $q$ corresponds to a small $r$.
Furthermore, as shown in (\ref{I_form}), $I(q,r)$ gets lower by $q^4$ in a high $q$ area.
Therefore, the number of event detections in the higher $q$ area became fewer in number, as shown in the SAS pattern graph.
As a result, the high $q$ area in the SAS pattern was noisy, and therefore, the small $r$ area in the inference results was also noisy.  
In fact, the IFT and ML results had a large amount of noise in a small radius area.
However, the noise is thought to have been reduced by the prior settings in the VB inference.

\begin{figure}[tb]
\centering
\includegraphics[width=0.3\textwidth]{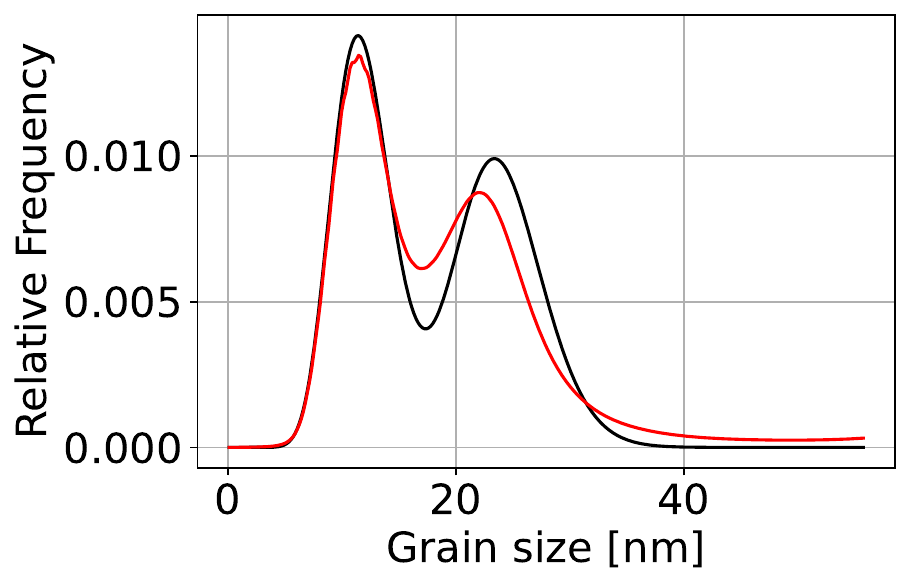}
\caption{IFT result from ideal SAS pattern}\label{idealiftexp}
\end{figure}

\begin{figure}[tb]
\centering
\includegraphics[width=0.3\textwidth]{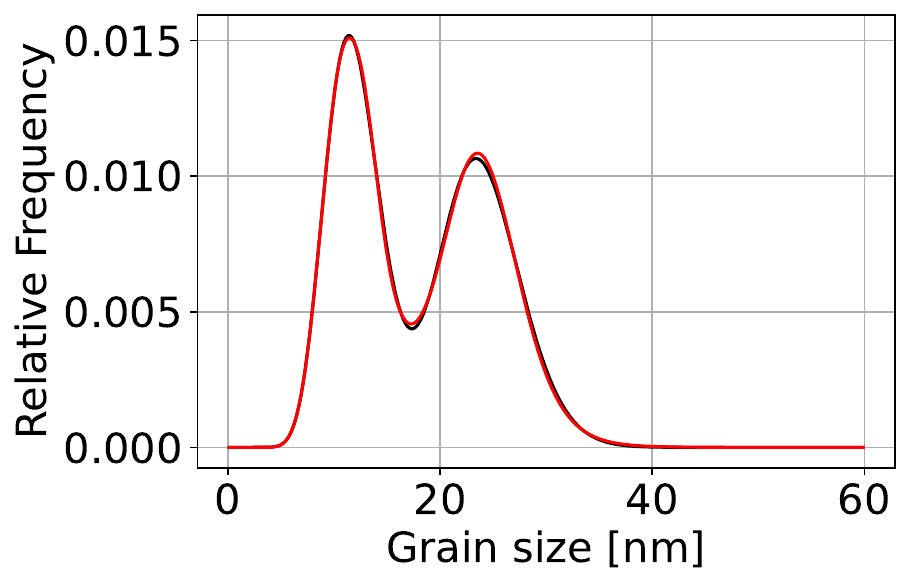}
\caption{ML result from ideal SAS pattern }\label{idealmlexp}
\end{figure}

\begin{figure}[tb]
\centering
\includegraphics[width=0.3\textwidth]{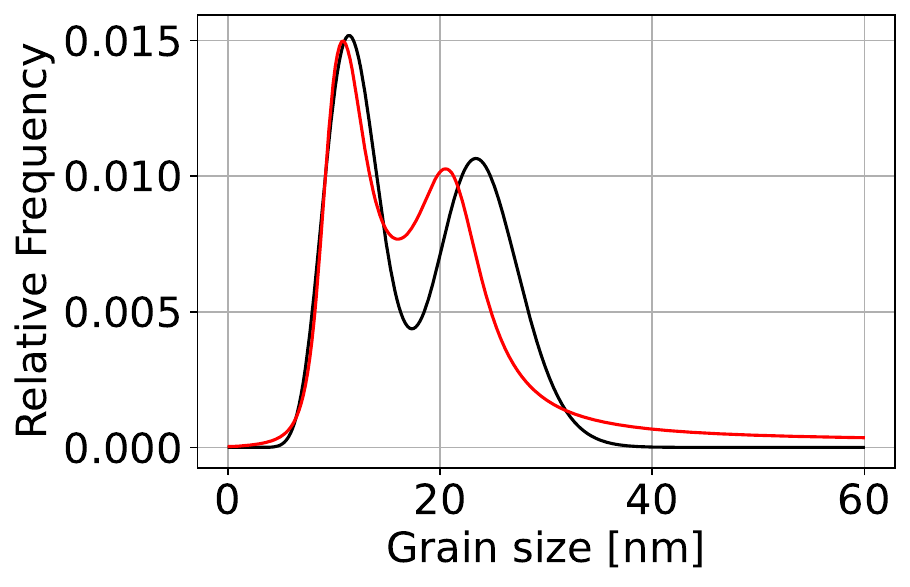}
\caption{VB result from ideal SAS pattern}\label{idealvbexp}
\end{figure}

To clarify such noise effects, Figures \ref{idealiftexp}, \ref{idealmlexp} and \ref{idealvbexp} show the results for the ideal SAS pattern, which is completely in proportion to $S(q)$ (50,000 events).
All of the results were very accurate. Though, in the large $r$ area, the VB was less accurate than the ML.
This shows that it is the better method to switch methods to perform the inverse estimation: the ML for a less-noisy SAS pattern and the VB for a noisy one.

\subsection{Experiment 2: EDS (Energy Dispersive X-ray Spectroscopy) }

\subsubsection{Overview}

Experiment 2 is related to EDS, which is a technology for performing compositional analysis by detecting characteristic X-rays generated by electron beam irradiation and then analyzing them spectroscopically by energy. The EDS is often implemented on the SEM (Scanning Electron Microscope).
In EDS, when an electron beam is irradiated onto an atom, the electrons of the atom are excited out. This allows electrons to enter the atom. X-rays are emitted during the transition, having a unique energy pattern by the element of the atom. By measuring them, it is possible to perform elemental analysis.
Although EDS pattern-shape-based analysis\cite{SOLE200763} is known, methods with handling the EDS data as multivariable are discussed here. 

\subsubsection{Experimental settings}

\begin{figure}[tb]
\centering
\includegraphics[width=0.35\textwidth]{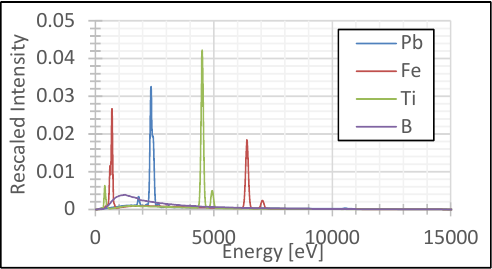}
\caption{Simple substance SMIs}\label{sssp}
\end{figure}

For Experiment 2, two types of EDS datasets were generated with simulation software DSTA-II\cite{DSTA}: a simple substance dataset and a compound dataset.

The simple substance dataset consists of the simulated spectral signals of 91 elements, such as B, C, Fe, Pb, Ag, U, Pu, and so on (that is, solid materials from B to Am in the periodic table).
Figure \ref{sssp} plots actual rescaled intensity signals of Pb, Fe, Ti, and B in the simple-substance dataset. The vertical axis indicates relative intensity, where summation of the signals is set to 1.0. The horizontal axis shows the energy, which is in proportion to the wavenumber $q$. This shows the spectral pattern characterizes the element.

\begin{figure}[tb]
\centering
\includegraphics[width=0.35\textwidth]{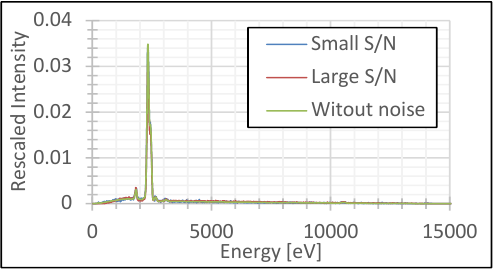}
\caption{Test data of PbS }\label{testsp}
\end{figure}

The compound dataset was generated for the test data, representing spectral data of 10 compounds consisting of less than 5 elements.
Three types of data were generated for each compound: the signal without noise, the signal with small noise, and the signal with large noise.
To make a large noise in terms of signal-to-noise ratio, the electric current is reduced when the large noise simulation was performed. 
Figure \ref{testsp} plots data from PbS. This shows that the noise is sufficiently smaller than the signal, and the elements are expected to be identified.
Note that the noiseless data has simulated fluctuations similar to Experiment1, and the noisy data was generated with additional noise. 

The process of identifying the components using this data can be considered as the inverse estimation of (\ref{S_cal}).
A substance of the simple substance dataset can be labeled by $r_i$, and the spectral signal of it can be handled as a vector along $\{q_k\}$.
That is, $r_i=\{$B, C, N, $\cdots$ Am$\}$, and the simple substance spectral signals can be written as a set of the $k$-components vector labeled with $i$.
Therefore, they form a matrix $I_{i,k} = I(q_k, r_i)$.
In addition, features of the compounds are handled as vectors.
The ratio of the substance in a compound is a vector $a_i = f(r_i)$, and the signals of the compounds consist of a vector $s_k=S(q_k)$ along $\{q_k\}$.
Because the signal of the compounds is the summation of the simple substance signals,
\begin{align}
S_k = \sum_i a_i I_{i,k}. \label{eds_s}
\end{align}
This formula is same as (\ref{S_cal}). Therefore, the ratio $a_i$ of the compound elements can be obtained with the proposed VB inference.
The elements in the compound are identified by selecting the substance with the large ratio. 

The datasets were processed with the proposed VB inference, and the SVD (based on the pseudoinverse matrix) and the ML inference\cite{asahara2020em} , were also applied to the datasets for comparison.
In the VB and ML inference, instead of checking convergence, 100 iterations were performed.
In the SVD inference, the pseudoinverse matrix of $I_{i,k}$ is calculated and multiplied by the signal $S_k$ of the compound.
According to (\ref{eds_s}), this leads to the inferred $a_i$.
The hyperparameters $\alpha_{0i}$ were set as $\alpha_{0i}=1$ in the VB inference.

The result of the VB, ML, and SVD inference is $a_i$.
Although the larger components of $a_i$ indicate the elements of the compounds, the number of the elements is not determined in this inference.
In Experiment 2, the number of elements is set equal to the truth. For example, if the target is PbS, the number of the elements is set to 2.
In the case that the inferred elements are completely equal to the truth, the inference is considered correct.

\subsubsection{Results}

\begin{table*}[h]
\centering \footnotesize
\caption{EDS inference results} \label{exp2_results}
\begin{tabular}{c|cccccccccc||c}
\hline
Material & PbS&MoS$ _2$&SrWO$ _4$&BaTiSi$ _3$O$ _9$&Ti(BaSi$_2$O$_5$)$_9$&TiO$_2$&Cr$_2$O$_3$&TiN&VN&Fe$_3$C&score\\
\hline
\hline
Truth & Pb,S&Mo,S&O,Sr,W&Ba,O,Si,Ti&Ba,O,Si,Ti&O,Ti&Cr,O&N,Ti&N,V&C,Fe\\
\hline
Without noise&\\
\hline
VB(proposed) &Pb,S&Mo,S&O,Sr,W&Ba,O,Si,Ti&Ba,O,Si,Ti&O,Ti&Cr,O&N,Ti&N,V&C,Fe&10/10\\
SVD &Ge,Th&Mo,S&Nd,Th,Yb&Cd,Ga,Ge,Nd&Cd,Ga,Ge,Nd&Cd,Th&Th,Yb&Ge,Th&Nd,Th&Nd,Th & 1/10\\
ML &Mo,S&Mo,S&O,Si,Sr&O,Rb,Si,Ti&O,Rb,Si,Ti&O,Ti&Cr,O&N,Ti&N,V&F,Fe & 5/10\\
\hline
Small noise&\\
\hline
VB(Proposed) &Pb,S&Mo,S&O,Sr,W&Ba,O,Si,Ti&Ba,O,Si,Ti&O,Ti&Cr,O&N,Ti&N,V&C,Fe& 10/10\\
SVD &Pb,Th&Mo,S&Nd,Th,Yb&Cd,Ga,Ge,Nd&Cd,Ga,Ge,Nd&Nd,Th&Th,Yb&Ge,Th&Nd,Th&Nd,Th& 1/10\\
ML &Mo,S&Mo,S&Si,Sr,W&O,Rb,Si,Ti&O,Rb,Si,Ti&O,Ti&Cr,O&N,Ti&N,V&F,Fe& 5/10\\
\hline
Large noise&\\
\hline
VB(Proposed) &Pb,S&Mo,S&O,Sr,W&Ba,O,Si,Ti&Ba,Si,Ta,Ti&O,Ti&Cr,O&Kr,Ti&N,V&C,Fe& 8/10\\
SVD &As,Pb&Nd,Th&Nd,Th,Yb&Cd,Ge,Nd,Yb&Ga,Ge,Nd,Th&Ge,Yb&Th,Yb&Nd,Th&Ge,Th&Fe,Ga & 0/10\\
ML &Mo,S&Mo,S&O,Si,Sr&O,Rb,Si,Ti&O,Rb,Si,Ti&O,Ti&Cr,O&Ba,Ti&N,V&F,Fe& 4/10\\
\hline
\end{tabular}
\end{table*}

The inference results are listed in Table \ref{exp2_results}.
Each column represents the inferred elements in the compound. The row 'Truth' raw at the top of the table gives the true elements.
By comparing these, the correct answer rates are calculated as the accuracy score. 

This result shows the VB inference performed outstandingly well. 
The VB inference gave perfect results on the noiseless and low-noise data, and even on the noisy data, there are only two errors.
In contrast, the SVD inference almost failed. 
The ML inference showed that half of the substances were correctly inferred.

In the results, the MoS$_2$ results are almost correct, and the Ti(BaSi$_2$O$_5$)$_9$ results are inaccurate.
These compounds are considered easy and difficult to infer, respectively.

\subsubsection{Discussion}

The SVD inference of EDS gave extremely low accuracy.
The reason is assumed to be that the SVD does not constrain the result to be positive.
In fact, the SVD inference results $a_i$ have about 50 of negative values. 

\begin{figure}[tb]
\centering
\includegraphics[width=0.35\textwidth]{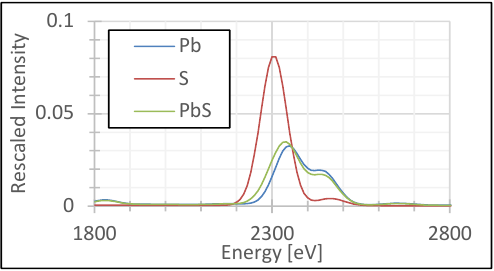}
\caption{Spectral comparison of Pb, S, and PbS}\label{subs}
\end{figure}

\begin{figure}[tb]
\centering
\includegraphics[width=0.35\textwidth]{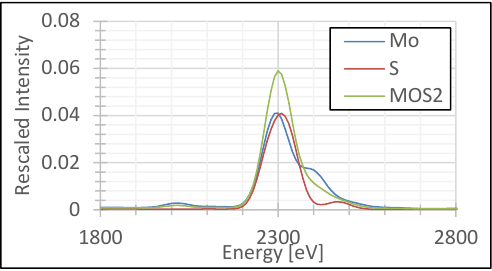}
\caption{Spectral comparison of Mo, S, and MoS$_2$}\label{mos}
\end{figure}

In the ML inference results of EDS, there are both correct and incorrect results.
To find out what makes this difference, Figure \ref{subs} and \ref{mos} shows close-up plots of SMI as examples.
The graph shown in Fig. \ref{subs} plots PbS, Pb and S, and the graph in Fig. \ref{mos} plots MoS$_2$, Mo and S.
Only the VB inference gave the correct results for PbS, while most of the results are correct for MoS$_2$.

The main difference between PbS and MoS$_2$ is the number of the peaks.
Both PbS and MoS$_2$ have a large peak around 2300eV, but only PbS has another peak at 2450eV.
Since S, which is included by both PbS and MoS$_2$, also has a peak at 2450eV,
the difference is caused by the difference between Pb and Mo.
The Pb has a peak at 2450eV similar to PbS, while the Mo has a peak at 2400 eV.
The MoS$_2$ has only one peak because the peak of Mo and S peaks cancel each other out.
For compounds with clear features such as Mo, it is assumed that the inference can be performed correctly using SVD and ML inference.
For the compounds composed of the elements with similar spectral shapes such as PbS,
however, the inference should be more accurate because the inference should be quantitively accurate even if the signals have fluctuations.
Since the VB inference is considered to be numerically accurate even with noise according to Experiment 1,
the VB inference can give correct results for the compounds such as PbS.

\section{Conclusion and Future Work}

The VB inference for SMI inverse estimation was proposed, and the experimental results were demonstrated in the two experiments of SAS and EDS. 
It was confirmed that the VB inference is accurate if the data has noise.
Therefore, it can be concluded that the VB inference can be helpful for SMI inverse estimation.

As future work, automatic determination of the components is needed in actual usecases.
For example, in EDS, several substances are included by the compound in general. However, the proposed method needs to input the number of the substances.
Another future work is to make nonlinear effects taken into account.
The proposed method assumes linear combination of wave components.
However, nonlinear effect like wave inference are utilized in measurements such as laser interferometer. 
The VB inference is needed to be extended for handling such nonlinear effect.

\bibliographystyle{unsrt}
\bibliography{sigproc}

\end{document}